\documentclass[3p]{elsarticle}
\usepackage{lineno,hyperref}
\usepackage{pdflscape}
\usepackage{pdfpages}
\usepackage{titletoc,tocloft}
\modulolinenumbers[5]

\usepackage{booktabs}
\usepackage{array}
\usepackage[gen]{eurosym}
\usepackage[longtable]{multirow}
\usepackage{color,soul}
\usepackage{footnote}
\usepackage{longtable}
\usepackage{threeparttable}
\usepackage{lscape}
\usepackage[usestackEOL]{stackengine}
\usepackage{placeins}
\usepackage{natbib}
\usepackage[inline, shortlabels]{enumitem}
\usepackage{amsmath,amsfonts,amsthm,amssymb,bm} 
\usepackage{fixmath}
\usepackage{setspace}
\usepackage{adjustbox}
\usepackage{dtklogos}
\usepackage{tikz}
\usepackage{diagbox}
\usepackage{makecell}
\usepackage{caption}
\usepackage{dsfont}
\usepackage{subcaption}
\usepackage{graphicx}
\usepackage{glossaries}
\usepackage{comment}
\usepackage{algorithm}
\usepackage{algpseudocode}
\usepackage{float}

\usepackage{adjustbox}
\usepackage{rotating}
\usepackage{breakcites}

\usepackage{tabularx}
\usepackage{xcolor}

\hypersetup{
	bookmarks=true,         
	unicode=false,          
	pdftoolbar=true,        
	pdfmenubar=true,        
	pdffitwindow=false,     
	pdfstartview={FitH},    
	pdftitle={My title},    
	pdfauthor={Author},     
	pdfsubject={Subject},   
	pdfcreator={Creator},   
	pdfproducer={Producer}, 
	pdfkeywords={keyword1} {key2} {key3}, 
	pdfnewwindow=true,      
	colorlinks=true,       
	linkcolor=red,          
	citecolor=blue,        
	filecolor=magenta,      
	urlcolor=blue}           
	
\newcolumntype{P}[1]{>{\centering\arraybackslash}p{#1}}
\newcolumntype{C}[1]{>{\centering\arraybackslash}m{#1}}
\newcolumntype{L}{>{\raggedright\arraybackslash}X}

\journal{Communications in Transportation Research}
\newcommand{\subsubsubsection}[1]{\paragraph{#1}\mbox{}\\}
\biboptions{authoryear}

\begin{document}
\begin{frontmatter}

\title{Development of a graph neural network surrogate for \\travel demand modelling} 

\author[TUMAddress]{Nikita Makarov}
\author[TUMAddress]{Santhanakrishnan Narayanan\corref{cor1}}
\ead{santhanakrishnan.narayanan@tum.de}
\author[TUMAddress]{Constantinos Antoniou}
\address[TUMAddress]{Technical University of Munich, Arcisstrasse 21, 80333 Munich, Germany}
\cortext[cor1]{Corresponding author}

\begin{abstract}
As urban environments grow, the modelling of transportation systems becomes increasingly complex. This paper advances the field of travel demand modelling by introducing advanced Graph Neural Network (GNN) architectures as surrogate models, addressing key limitations of previous approaches. Building on prior work with Graph Convolutional Networks (GCNs), we introduce GATv3, a new Graph Attention Network (GAT) variant that mitigates over-smoothing through residual connections, enabling deeper and more expressive architectures. Additionally, we propose a fine-grained classification framework that improves predictive stability while achieving numerical precision comparable to regression, offering a more interpretable and efficient alternative. To enhance model performance, we develop a synthetic data generation strategy, which expands the augmented training dataset without overfitting. Our experiments demonstrate that GATv3 significantly improves classification performance, while the GCN model shows unexpected dominance in fine-grained classification when supplemented with additional training data. The results  highlight the advantages of fine-grained classification over regression for travel demand modelling tasks and reveal new challenges in extending GAT-based architectures to complex transport scenarios. Notably, GATv3 appears well-suited for classification-based transportation applications, such as section control and congestion warning systems, which require a higher degree of differentiation among neighboring links. These findings contribute to refining GNN-based surrogates, offering new possibilities for applying GATv3 and fine-grained classification in broader transportation challenges.
\end{abstract}

\end{frontmatter}

\section{Introduction}
\label{sec:Intro}
Transportation plays a critical role in today's society. The modern urban landscape is characterized by ever-increasing mobility demands, necessitating sophisticated and efficient transport systems. Focusing on long term planning, travel demand modelling plays a pivotal role in shaping the future, to meet the needs of growing populations and evolving societal demands. However, the complexity of urban environments, with their intricate networks of roads, public transit systems, and diverse travel patterns, presents a formidable challenge in accurately modelling and simulating transportation systems for the future. The need for scalable, adaptable, and computationally efficient modeling techniques has therefore become more pressing than ever.

In recent years, the development of surrogate models has emerged as a promising approach to tackle the complexity of strategic transport planning. Unlike traditional models that often require extensive computational resources and expert calibration, surrogate models leverage data-driven methodologies to streamline the evaluation of complex mobility solutions. These models have demonstrated their potential in various domains by not only reducing computational costs but also enhancing predictive accuracy (e.g., \citealp{Choi.09052023}). Moreover, such models can also be tuned directly using data and auto-calibration procedures are available. User-friendly models like these that can evaluate complex mobility options are being expected much by the city authorities \citep{Narayanan.2023d}. 

A proof-of-concept by \citet{Narayanan.2024} demonstrated the feasibility of using Graph Convolutional Networks (GCNs) as surrogates for travel demand modelling. This foundational work explored the potential of Graph Neural Networks (GNNs) in replicating the traditional four-step model, offering a pathway toward more sophisticated applications such as agent-based models. While these initial findings were promising, key challenges remain unaddressed - particularly concerning model expressiveness and stability in prediction accuracy. This paper builds upon and significantly advances the previous work by introducing novel methodologies that enhance both the depth and applicability of GNN-based surrogate modeling in strategic transport planning. Specifically, we introduce: 

\begin{itemize} 
    \item GATv3: A Graph Attention Network (GAT) variant that overcomes the depth limitations of existing GAT models caused by over-smoothing. By incorporating residual connections, GATv3 enables deeper and more expressive network architectures.
    \item Synthetic data augmentation strategy: To address data scarcity and improve model performance, we introduce a synthetic data generator that effectively expands the augmented training dataset from \citealp{Narayanan.2024}. Purely synthetic data is the one generated procedurally, whilst augmented implies that a true data source is employed, but modified to produce more data. This new approach is implemented to enrich the model's learning capabilities without inducing overfitting. 
    \item Fine-grained classification framework: We propose an approach that transforms traditional regression problems into a fine-grained classification setup. This methodology not only ensures numerical precision comparable to regression but also enhances model stability, thereby making it more suitable for strategic transport applications.  
\end{itemize}

The remainder of this paper is structured as follows: Section \ref{sec:LitRev} provides a brief review of the prior work. Section \ref{sec:formulation} elaborates on the methodology, detailing the architectures of GAT and the proposed variant, synthetic data generation procedure and the different experimental setups. The experimental results are presented in Section \ref{sec:Results}. Finally, Section \ref{sec:conclusions} offers concluding remarks.

\section{Literature review}
\label{sec:LitRev}
This section focuses on the existing literature related to the application of (i) GNNs and (ii) surrogate models for transport planning. The reviewed studies were gathered through the Scopus database using an \href{https://github.com/nsanthanakrishnan/Scopus-Query}{open source python script}\footnote{\href{https://github.com/nsanthanakrishnan/Scopus-Query}{https://github.com/nsanthanakrishnan/Scopus-Query}} from \citet{Narayanan.2022h}.

\subsection{GNNs for transport planning}
\label{litrev:dl}
GNNs are slowly gaining significant traction in the domain of transport planning due to their ability to capture intricate spatial dependencies within complex urban networks. For example, GNNs are used to predict the demand for micromobility modes. In this direction, \citet{Lin.2018} proposed Graph Convolutional Neural (GCN) Network model to forecast station-level hourly demand for a bike-sharing system in New York City and conclude that such models show promising results. \citet{Ding.2022} utilised a GNN model to study the effect of a dockless bike-sharing scheme on the the demand of the London Cycle Hire scheme. For a shared E-scooter system, \citet{Song.2023} developed a GNN model to predict the demand in Louisville, USA. Both of these studies also conclude that GNN models make better predictions.

GNNs are also used for studying metro and taxi trips. For example, \citet{Zhao2023} designed an adaptive GCN model to predict short-term passenger flow for metro systems. \citet{Wang.2024} developed a spatio-temporal GNN model for taxi route assignment. \citet{Tygesen.2023} conduct experiments on NYC Yellow Taxi and provide insights on the kinds of connections GNNs use for spatio-temporal predictions in the transport domain. They also perform tests for short term traffic prediction. Similarly, \citet{Cui.2020} incorporate graph wavelet and employ gated recurrent structure for network-wide traffic forecasting. \citet{Xu.2023} conclude that graph based models perform better than other machine learning models for network-wide short-term traffic speed prediction and imputation. Graph based neural network models are also used for real-time incident prediction (e.g.,  \citealp{Tran.2023}) and can be helpful in solving the air transportation nexus among safety, efficiency and resilience \citep{Wandelt.2024}.

\subsection{Surrogate modelling in the field of transportation}
Surrogate modelling is a macro-modeling technique to approximate expensive functions and to replicate input–output relationships.  Surrogate models are also called meta- or response surface models and posses beneficial qualities, such as cheaper evaluation costs and sometimes, better accuracy. To approximate complex simulation systems, deep learning based surrogate models are appropriate, as they are suitable to represent non-linear relationships and can handle high dimensional problems well. In the field of traffic forecasting, \citet{Vlahogianni.2015} introduced the concept of surrogate modelling and suggests some approaches to specific optimization problems that a transportation researcher may face when dealing with forecasting models. The approaches can be adapted to various transportation problems and the results convey that surrogate models can generate more accurate predictions in significantly reduced times. 

\citet{Jariyasunant2014} developed a computation surrogate for travel feedback. They state that such surrogates can help to overcome scaling issues. Focusing on dynamic user equilibrium, \citet{Ge.2020} formulated a dynamic user equilibrium traffic assignment model incorporating surrogate dynamic network loading models. They conclude that the usage of surrogate approaches reduces the computational times by up to 90\%, while at the same time, leading to lower errors. Surrogate models are also used to assign Dedicated Bus Lanes (DBLs) across a large-scale road network. The results of numerical studies from \citet{Li.2022} show that surrogate based methods for optimal DBL allocation scheme can save a substantial amount of time, especially with limited computational resource, while also leading to around 5\% better network performance than the costly simulation approaches. 

The concept of GNN based surrogate modelling in the context of travel demand modelling emerged very recently, with the pioneering work of \citet{Narayanan.2024}. They laid a foundational groundwork by investigating the feasibility of developing a GCN based surrogate for strategic transport planning. Their study delve into both classification and regression setups, and based their analysis by generating data from Greater Munich metropolitan region. The experimental results show that GNNs have the potential to act
as transport planning surrogates and the deeper models perform better than their shallow counterparts. However, the models suffer performance degradation with an increase in network size. 

\subsection{Summary and research gaps}
In this section, a summary of the reviewed studies and the identified gaps are provided, which act as a motivation for the research
objectives of this study. On the one hand, the existing literature on the application of GNNs shows that they have the potential to be utilised for transport modelling. On the other hand, surrogate models are found to be computationally effective and generate more accurate outputs. Therefore, it is advantageous to develop a GNN surrogate for strategic transport planning, the foundation for which has already been laid by \citet{Narayanan.2024}.

The study of \citet{Narayanan.2024} demonstrated the efficacy of GCNs and showcased how GCNs outperformed traditional machine learning models by effectively leveraging the graph structure inherent in transportation networks. Nevertheless, the limitations of GCNs (e.g., the neighbourhood aggregation function used in the GCN puts equal weight on all neighbours, while all neighbours may not be equally relevant) and that of the classification and regression setups necessitate the investigation of the use of advanced GNN architectures and the analysis of a new modelling setup, which will be the focus of this study.

\section{Methodological framework}
\label{sec:formulation}
\begin{figure}[!ht]
  \centering
  \includegraphics[scale=0.24]{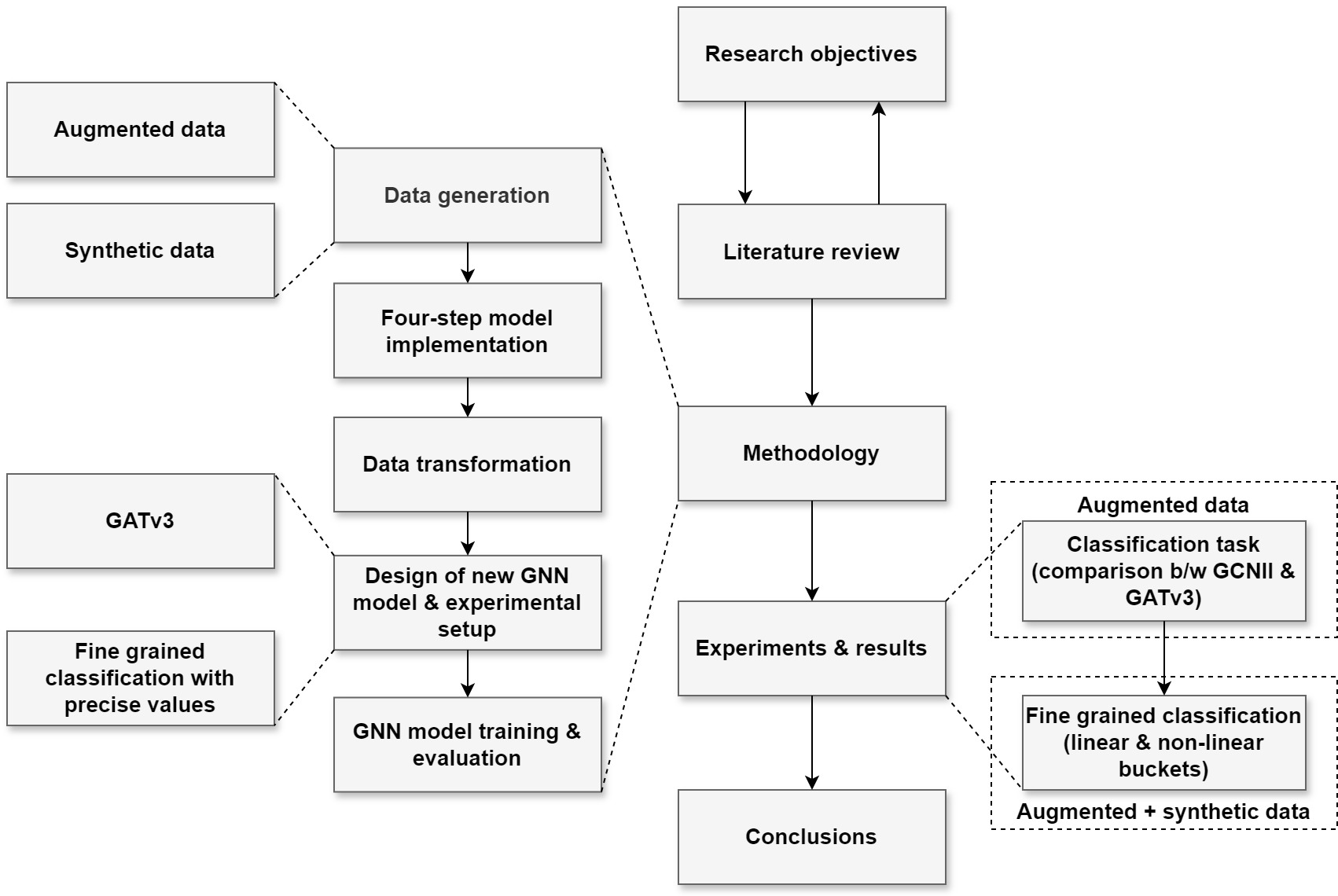}
  \caption{Methodological framework}
  \label{fig:Methodology}
\end{figure}

This study builds over the methodological framework established in \citet{Narayanan.2024}, as shown in Figure \ref{fig:Methodology}. Therefore, the procedural sequence is as follows: (i) data generation for the four-step model, (ii) simulation in the four-step model, (iii)  data transformation (i.e., direct graph generation) and the creation of the training data (iv) GNN model training and model evaluation. The regularization methods (i.e., Dropout and GraphNorm), loss functions (cross entropy for the classification tasks and mean squared error for the regression tasks) and evaluation metrics (accuracy and F1-score for the classification tasks and mean absolute error and R\textsuperscript{2}) from there are also utilised in this study. For detailed information of the setup, the reader is referred to the original study. 

As an addition to the existing setup, this study develops and investigates further GNN model arcitectures and additional data generation process, which are introduced in the following sections. Specifically, we describe the newly proposed GATv3 model, synthetic data generation and fine grained classification approach with the ability to obtain precise values as in the case of classical regression. \citet{Narayanan.2024} implemented augmented data generation based on the network of Munich metropolitan region, but this study utilises the augmented data as well as synthetic data. With regards to the difference between the two, purely synthetic data is the one generated procedurally, whilst augmented implies that a true data source is employed, but modified to produce more data. 

The implementation of the entire pipeline is performed in Python. The baseline models are estimated using Scikit-Learn \citep{scikit-learn.2011}.  GNN models are implemented in PyTorch Geometric \citep{fey.2019}, which is an extension to the PyTorch library \citep{wallach.2019}. The entire code (including the ones for data generation and processing) is provided in a private GitHub repository, which is found at \url{https://github.com/nikita68/4_step_model_surrogate}. Access to it is possible upon request. Nevertheless, the processed datasets are separately available at \url{https://github.com/nikita6187/TransportPlanningDataset}.

\subsection{Graph attention networks}
\citet{Narayanan.2024} developed a GNN surrogate based on Graph Convolutions Networks (GCN). The neighbourhood aggregation function used in the GCN puts equal weight on all neighbours. However, not all neighbours may be equally relevant. An example would be an incoming road having a high traffic demand, whilst another having zero at a junction. Thus, the information relevant for the outgoing road comes more from the high demand incoming road. Therefore, in this section, the graph attention networks are presented, together with an extension to enable them to generalize to deeper networks. 

To be able to understand the difference between different neighbours, the concept of attention was introduced to form the Graph Attention Networks (GAT) \citep{brody.2021}. Attention is determined by calculating a set of weights for each neighbour of a node, with every weight being between 0 and 1, and all the weights summing up to 1. When summing up the transformed feature vectors of a node's neighbours, the vectors are multiplied by the respective weights. The weights themselves are calculated by applying a one layer MLP, which takes in both the original node's and its neighbour's feature vector as input. Please note that the GAT model estimated in this study is GATv2 \citep{velivckovic.2017,  brody.2021}. 

GATv2 still suffer from the over-smoothing problem, much like the original GCN. This results in both models performing poorly when trying to stack more than 5 layers. To the best of the authors' knowledge, there is no solution available to this problem at the time of this study. Inspired by the results of GCNII, an extension is proposed to use the same residual trick from \citet{chen.2020}, which adds states from a lower layer directly to a higher layer and avoids the distinct local features getting vanished in the higher layers. Thus, the GATv2 formulation becomes as follows:

\begin{align}
\label{models:gatv3:formula}
\begin{split}
    {\mathbold{n}}_i^j =& \, \sigma\left( (1 - \alpha) \cdot \left( \sum_{k  \in n(i) \cup \{i\}} \alpha_{ik} \cdot {\mathbold{n}}_k^{j-1} \times \mathbold{W}^j \right) + \alpha \cdot {\mathbold{n}}_i^0 \right), \\[0.8ex]
    & \alpha_{ik} = \, \sigma_{softmax} \left( N_{ff} \left( \left[ {\mathbold{n}}_k^{j-1};{\mathbold{n}}_i^{j-1} \right] \right) \right)_k, \\[0.8ex]
    & \text{with } \sum_{k} \alpha_{ik} = 1, \text{ and } \alpha_{ik} \in [0, 1].
\end{split}
\end{align}
where,\\
$\alpha$ is a fixed hyperparameter \\
$\alpha_{ik}$ is the attention weight of node $i$ for neighbour $k$, with the attention weights summing up to 1 for all of a given nodes neighbours, and each weight being between 0 and 1 \\
$\sigma$ is the activation function \\
$\sigma_{softmax}$ is the activation function that forces all outputs to be between 0 and 1, and normalizes all outputs so that they sum to 1 \citep{nwankpa.2018}\\
$k$ represents all the direct neighbours of node $i$ \\
$N_{ff}$ is a one layer MLP \\
${\mathbold{n}}_i^0$ is from the input layer for node $i$ \\ 
${\mathbold{n}}_i^j$ is the feature of node $i$ in GAT layer $j$ \\
$\mathbold{W}^j$ is the weight matrix of layer $j$ \\ 
$[;]$ operator concatenates the input vectors  \\

Residual connections were originally proposed in the field of computer vision to mitigate vanishing gradient problems. In the current case, even though it is a minor change performed on the GATv2 model, it is an effective solution. It enables the possibility of developing deeper GAT models, as will be shown in later sections. From now, this model is denoted as $GATv3$. Note that, due to the extra computations, the number of GAT layers which can fit onto one GPU is significantly less than those of GCNs.

\subsection{Synthetic data generation}
\label{ssec:SyntheticData}
For data generation, \citet{Narayanan.2024} implemented augmented data generation, wherein the network of Munich metropolitan region is used to generate multiple subnetworks for model training. It is often seen that more data significantly helps deep learning models \citep{wang.2017}. However, the original network data source utilised for the augmented dataset in \citet{Narayanan.2024} has only a finite number of possible subnetworks, until the training dataset has a critical overlap with the validation and test datasets.  To overcome this issue, while still providing arbitrarily more data, this study introduces an additional data generation procedure, namely the synthetic data generation. 

A synthetic region generation algorithm is implemented, which does not use any original network as a source. This algorithm is motivated based on the works of \citet{parish.2001, smelik.2014, cogo.2019}. Assuming that the target amount of nodes ($\mathds{G}_{nodes}$) is provided as input, along with the components for generating zones and zonal information 
(as identical to that of the augmented data generation procedure), the algorithm performs the following steps:

\begin{enumerate}
    \item Create the required number of random nodes $\mathds{G}_{nodes}$ within a square region.
    \item Calculate the number of zones to be created, i.e., $\mathds{G}_{zones} = \mathds{G}_{nodes} * \mathds{G}_{zones/nodes}$. Select $\mathds{G}_{zones}$ among $\mathds{G}_{nodes}$, which are furthest away from each other to become zonal nodes.
    \item Generate zonal information on the previously selected (zonal) nodes.
    \item Generate routes between the zones in two steps.
    \begin{enumerate}
        \item Create $\mathds{G}_{zones}$ number of routes. Each route is created by randomly selecting two zones based on their number of residents and employees. Next, a simple heuristic algorithm starts at a node, examines the closest 7 neighbours, and selects the neighbour which is closest to the target zone as the next step along the route. This is repeated until the route is finished.
        \item Next, if any zone is not yet connected to the network, apply the above routing algorithm to the closest node part of the existing network.
    \end{enumerate}
    \item Delete the nodes that are not connected and split long segments of the network until target number of nodes is reached.
    \item Run the four step model on the data, and save the data and the output of the four step model.
\end{enumerate}

The synthetic dataset is generated with 10000 samples, with a target number of nodes between 15 and 80. Note, the synthetic dataset is used in this paper only as supplementary training data for the augmented dataset with further details being provided in the experiments section (Section \ref{sec:Models}).

\subsection{Experimental setup}
\label{sec:Models}
This section will focus on two distinct setups, i.e., (i) an initial classification setup to compare the different GNNs and (iii) a fine-grained classification setup to exploit the stability of classification, which also includes the proposal of a small transformation procedure to convert the classification results into actual traffic values. Finally, the impact of additional data will be explored using the fine-grained classification setup. 

\subsubsection{classification task (comparison between GCNII, GATv2 and GATv3)}
\label{sssec:comparisonModels}
The first experiment is focused on comparing the different GNNs. As \citet{Narayanan.2024} concluded that the classification problem (or more formally cross entropy) is more stable than the regression loss (i.e., MSE) with respect to the hyperparameters, this comparison will be based on the classification task. The real valued traffic data is inserted into a bucket, with the bucket index then being the target, which is also inspired in part by the levels found in congestion warning systems \citep{gerstenberger.2018}. The buckets used are  $[0 \, veh/h, 10\, veh/h)$, $[10 \, veh/h, 500 \, veh/h)$, $[500 \, veh/h, \infty \, veh/h)$.  These three buckets ensure that each bucket has enough samples, with a 62\%, 21\% and 17\% distribution respectively on the validation dataset. The buckets correspond to no transport usage, low transport usage and medium/high transport usage.

The augmented dataset is used. All the models in this experiment are evaluated using the average F1 score. For the GCNII, a grid search is performed over the number of GCNII layers $\in$ (5, 10, 50, 70), the size of each layer $\in$ (64, 256, 512), $\alpha$ $\in$ (0.1, 0.4) and $\theta$ $\in$ (0.5, 1.5). 
The best performing model on the validation dataset is based on 70 layers, each layer containing 512 hidden units, $\alpha$ = 0.1 and $\theta$ = 1.5. If not stated otherwise, the same parameters are used for all future GCNII experiments. This experiment further investigates whether the more expressive GAT models improve the predictions. In the preliminary analysis, it was observed that the GATv2 model had convergence problems when attempting to stack more than 5 layers, which is also supported by the observations in the literature \citep{chen.2020}. Due to the average graph diameter of the dataset being 36, it is mathematically impossible for a shallow GATv2 model to correctly capture the underlying travel patterns. Thus, a GATv3 grid search is performed, with either 20 hidden layers and 512 hidden units or 40 hidden layers with 256 hidden units. 
Based on the validation dataset, the best configuration is 20 hidden layers with 512 hidden units. 

\subsubsection{Fine grained classification}
\label{ssec:fineGrainedSetup}
\subsubsubsection{Overview of the setup}
To further exploit the stability of classification, this experiment is conducted to explore fine grained classification. Furthermore, a method is proposed to convert the classification results into real values. Comparing the F1 score does not make sense when using different buckets, as the the comparison becomes skewed toward the model configurations with fewer buckets. Thus, it makes more sense to use the following metrics: $R^2_{\, \ge10}$ and $MAE_{\, \ge10}$. However, in classification, only the probabilities are provided as outputs of the model. To convert the probabilities of buckets into a real value, a small transformation procedure is proposed using the expectation operator from statistics. Initially, a uniform distribution across each bucket is assumed and the expectation is taken, resulting in the mean value of the bucket. Then, the expectation over all buckets is applied, using the output of the model as the discrete probability distribution. The result is that the model outputs real values in veh/h. Formally,

\begin{equation}
    g(x)'_i = \mathds{E}_{g(x)_i} \left [ \mathds{E}_{uniform} \left[ b_k \right] \right] = \sum_k g(x)_i^k \cdot \frac{( b_{k,1} + b_{k,2})}{2}.
\end{equation}
where,\\
$g(x)'_i $ is the resulting output of the model for the current link $i$\\
$\mathds{E}_{g(x)_i}$ is the expectation using the probabilities provided by the GNN model\\
$\mathds{E}_{uniform}$ is the expectation over the bucket $k$, assuming a uniform distribution\\
$g(x)_i^k$ is the probability from the GNN model for the current link $i$ for bucket k\\
$b_{k,1}$ is the lower bound of the bucket whilst $b_{k,2}$ is the upper bound\\


\subsubsubsection{Optimal Bucket Strategy}
At first, an optimal bucket strategy needs to be found. A grid search is performed together with a detailed analysis of the best models. The input data is identical to the previous experiments, and the target data is always the bucket index for the classification setup. Two bucketing strategies are conducted: equidistant and non-linear bucketing. For equidistant buckets, the following configurations are explored, as they require little reconfiguration:
\begin{itemize}
    \item \textit{Buckets-e23} - 23 buckets in step size of 200 veh/h
    \item \textit{Buckets-e45} - 45 buckets in step size of 100 veh/h
    \item \textit{Buckets-e90} - 90 buckets in step size of 50 veh/h
    \item \textit{Buckets-e180} - 180 buckets in step size of 25 veh/h
    \item \textit{Buckets-e450} - 450 buckets in step size of 10 veh/h
\end{itemize}

When looking at the equidistant buckets, it is seen that many of the higher value buckets have no or few training samples, whilst lower values have many thousands of samples. The difference in sample sizes can greatly impact the performance of GNNs, and especially restrict their generalization behaviour. To compensate for this effect, non-linear bucket sizes are also explored, to ensure that all buckets have at least a few training samples. The following configurations are examined:

\begin{itemize}
    \item \textit{Buckets-nl38} - Steps of 25 from 0 veh/h to 250 veh/h, steps of 50 from 250 veh/h to 1000 veh/h, steps of 100 from 1000 veh/h to 1500 veh/h, steps of 500 from 1500 veh/h
    \item \textit{Buckets-nl54} - Steps of 25 from 0 veh/h to 500 veh/h, steps of 50 from 500 veh/h to 1500 veh/h, steps of 100 from 1500 veh/h to 2000 veh/h, steps of 500 from 2000 veh/h
\end{itemize}

Further configurations are not explored as the results implied that there is no significant further improvement. 

\subsubsubsection{Use of additional synthetic data}
Synthetic data to increase the training dataset has been shown in existing literature to improve deep learning models \citep{Nikolenko.2021}. The synthetic dataset is used only to expand the training data, with the validation and test datasets being kept the same as in the previous experiments to ensure proper comparisons. The new training dataset size is 
more than 2.5 times larger than the original dataset. The two previously best performing bucket configurations (nl54 and e90), as will be observed in Section \ref{sec:Results}, are then trained from scratch using the additional synthetic training data. Each training run of a model takes around 12-15 hours. 


The final experiments examine how GATv3 model perform using the training data supplemented by the synthetic samples. The data setup is identical to the previous experiment, but only using the \textit{Buckets-nl54} configuration. 
The GCNII is used with 70 layers, 512 hidden units, $\alpha=0.1$, $\theta=1.5$, 3 residual feed forward layers, GraphNorm after each convolution operation and a dropout of 25\%. The GATv3 model contains 20 layers each with 512 hidden units, based on 2 attention heads performing the averaging merging strategy, 3 residual feed forward layer, GraphNorm and dropout with 25\%.

\section{Experimental results}
\label{sec:Results}
The experiments described in the previous section are performed sequentially and the respective results are summarised in this section.

\subsection{Evaluation of different GNN architectures (GCNII and GATv3)}
Different GNN models are compared to find the optimal model setup and a classification task is used for this. The choice of classification as the primary task for comparing different GNN architectures was driven by methodological and practical considerations. Classification provides a simpler and more controlled experimental setting, allowing for a direct comparison of architectures without the added complexities of continuous target prediction. Moreover, classification tasks are generally more stable in terms of optimization and convergence, reducing the risk of confounding factors such as sensitivity to hyperparameters, which are more prominent in regression. Starting with classification also allowed for a structured experimental progression - by first assessing how well each architecture performs on a simpler task, we could identify potential strengths and weaknesses before moving to the more complex problem. Additionally, classification outputs are inherently more interpretable, enabling easier diagnostic analysis to understand model behavior before tackling a more nuanced regression setting. Overall, choosing classification as the initial task ensured a robust and systematic comparison of GNN architectures, minimizing external sources of variability and establishing a solid basis for the subsequent experiments.

In a preliminary analysis, it was observed that the GATv2 model had convergence problems when attempting to stack more than 5 layers, which is also supported by the observations in the literature \citep{chen.2020}. Due to the average graph diameter of the dataset being 36, it is mathematically impossible for a shallow GATv2 model to correctly capture the underlying travel patterns. Hence, the results are shown only for GCNII and GATv3 GNN models. The F1 score of the models can be found in Table \ref{results:q1:advanced_models:results}. It is observed that the GCNII model performs significantly better than all baselines. Note that the accuracy of the model is at 85\%, and most errors are observed in classifying the [10 veh/h, 500 veh/h) bin. For the bin of [0 veh/h, 10 veh/h), an accuracy of 95\% is achieved. On a different note, the configuration containing both the largest number of layers and largest layer size in the GCNII performs best. Nevertheless, the best performing model is the GATv3.

\begin{table}[h]
\centering
\setstretch{0.9}
\captionsetup{justification=centering}
\caption{F1 score corresponding to the test dataset, for the classification task comparing the different GNN models}
\label{results:q1:advanced_models:results}
\vspace{0.1cm}
\begin{tabular}{@{}lcc@{}}
\toprule
Model \hspace{2cm}              & $F1$ Score \\ \midrule
Majority Classifier & 0.25     \\
Random Forest       & 0.54    \\
MLP                 & 0.53     \\
GCNII               & 0.85    \\ 
GATv3               &  \textbf{0.87}   \\ \bottomrule
\end{tabular}
\end{table}

The convergence of all GNN models based on the direct graph is observed to be stable and level out at around 300 epochs. If not otherwise stated, all the presented GNN models in this study have similar convergence behaviour on the training and validation data. It can be concluded that a larger number of layers and thus more trainable parameters are helpful. Additionally, it is observed that all models perform similarly between their test and validation scores, verifying that the generalization of the models is highly probable. 

\subsection{Evaluation of fine grained classification}
In this section, the results of the classification performed on fine grained buckets are presented. By applying the double expectation operator, it is possible to convert probabilities to real values, with all details about the setup found in Section \ref{ssec:fineGrainedSetup}. Different bucket strategies reach good results, when compared to the baselines. However, only GCNII - Buckets-e90 slightly surpasses the $MAE_{\, \ge10}$ score of the best regression model. Between the best models of the linear and non-linear bucketing, the latter does not surpass the former. 

\begin{figure}[!htb]
    \centering
    \includegraphics[clip, trim=0cm 0cm 0cm 0cm, width=0.7\textwidth]{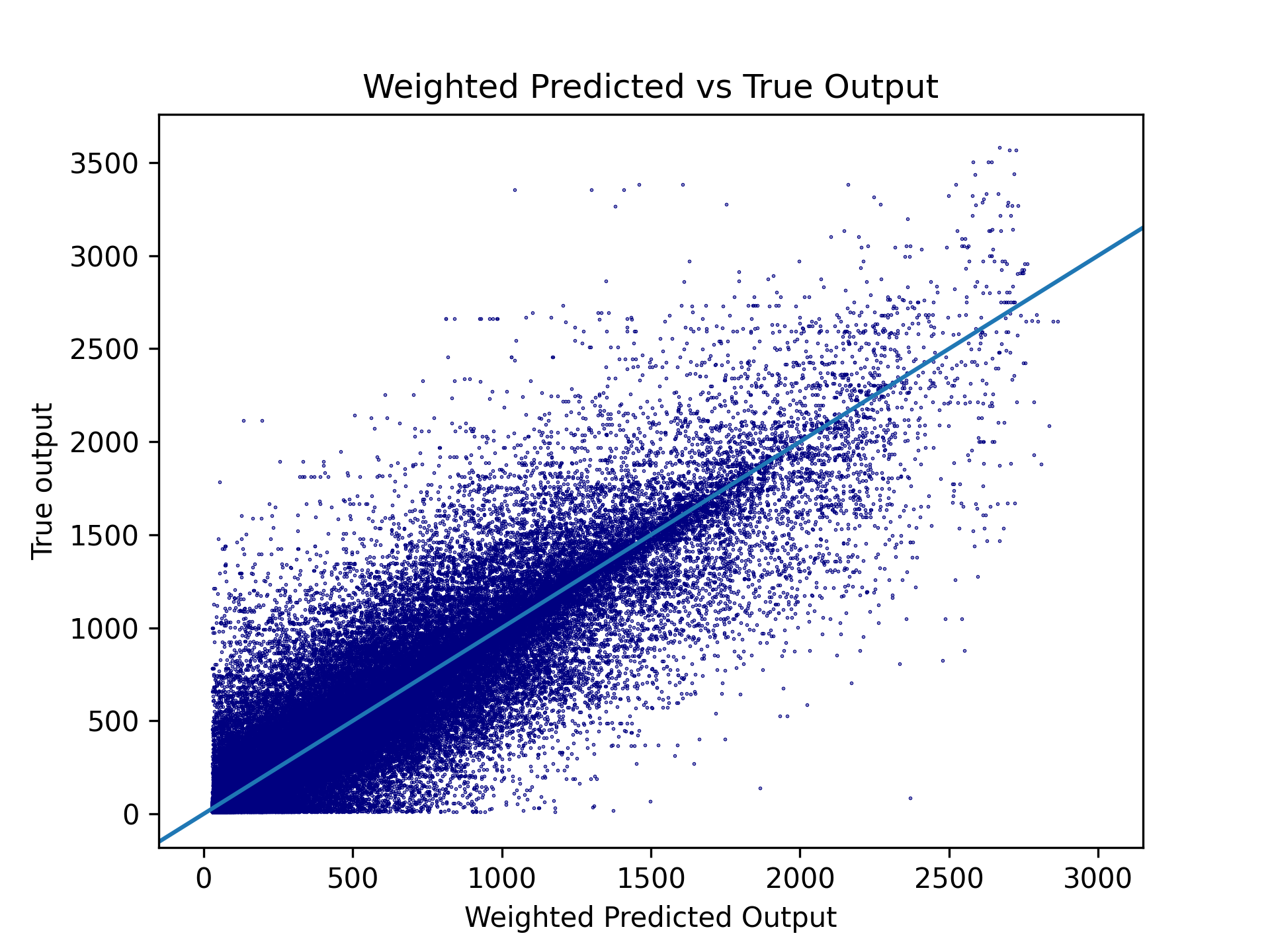}
    \captionsetup{justification=centering}
    \caption{Predicted output vs true output for the GCNII - Buckets-e90 - extra data model on the test dataset. Each point is a prediction for a single link, with both values being in veh/h. An optimal prediction would be along the light blue line. Notice that for true values above 2500, the model constantly predicts around 2600.}
    \label{fig:e90Pred}
\end{figure}

\begin{figure}[!htb]
    \centering
    \includegraphics[clip, trim=0cm 0cm 0cm 0cm, width=0.7\textwidth]{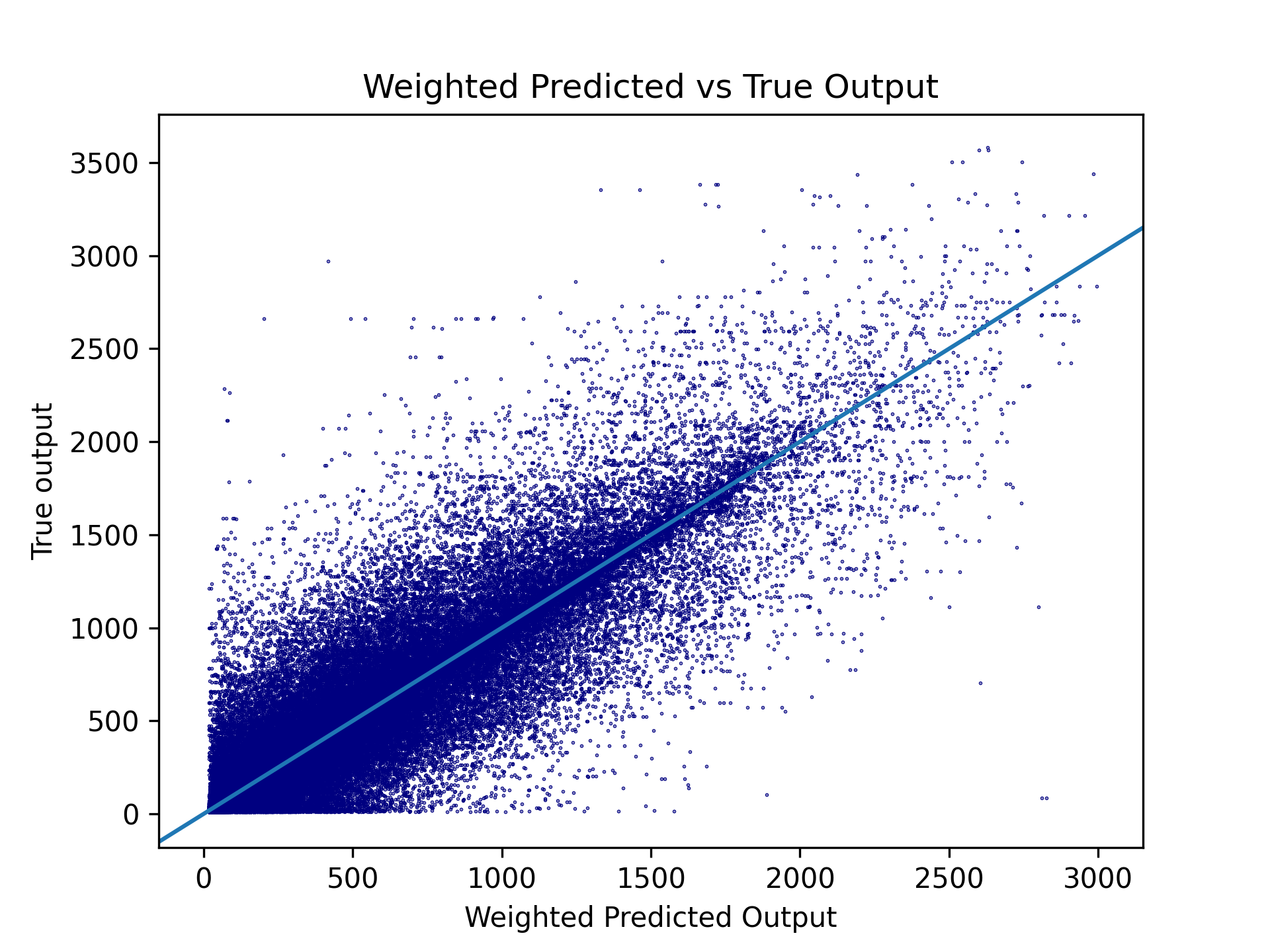}
    \captionsetup{justification=centering}
    \caption{Predicted output vs true output corresponding to the test dataset  for the GCNII - Buckets-nl54 - extra data model. Each point is a prediction for a single link, with both values being in veh/h. An optimal predicted would be along the light blue line. Observe that for true values over 2500, the model still predicts proportionally higher output values.}
    \label{fig:nl54Pred}
\end{figure}

The two best performing bucket configurations, nl54 and e90, are then trained from scratch using additional synthetic training data. Both the models with the extra data perform significantly better than the baselines and have an 16\% better performance than the models without the extra data. However, the best non-linear bucketing model has a slightly better performance than the best linear bucketing model, which is in contrast to the ones without the extra data. Looking more closely at the differences between the models ``GCNII - Buckets-e90 - extra data" and ``GCNII - Buckets-nl54 - extra data", it is found that the former predicts high values poorly. As seen in Figure \ref{fig:e90Pred}, all true values above around 2500 veh/h get assigned values around 2600 veh/h. On the other hand, the model with non-linear 54 buckets continues to predict proportionally high values, as shown in Figure \ref{fig:nl54Pred}. The effect comes probably from the high number of classes having few data points for training in the case of equidistant bucketing.

Compared to the regression setting, the fine grained classification returned more consistent results, probably as fewer steps are required. Such behaviour is especially desirable as it requires less configuration searching and reduces the overall model development time. Furthermore, extra synthetic data has shown to make a significant improvement in the performance of the models. The evaluation metrics of the experiment runs can be found in Table \ref{results:q3:extra_data_put:results}. The best performing model is the GCNII with extra data, and the performance is comparatively poor with GATv3.

\begin{figure}[!htb]
    \centering
    \includegraphics[clip, trim=0cm 0cm 0cm 1.35cm, width=0.78\textwidth]{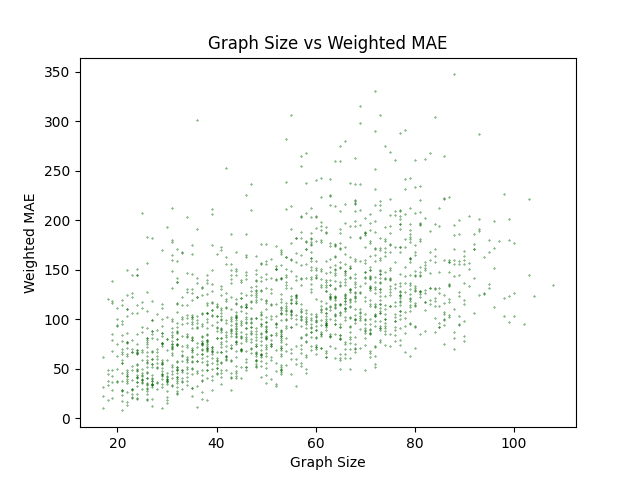}
    \captionsetup{justification=centering}
    \caption[$MAE_{\, \ge10}$ with respect to graph size.]{$MAE_{\, \ge10}$ with respect to graph size, i.e. number of nodes for the trained GNN model on the test dataset. Each point is the average absolute error of a single graph, with the errors coming from incorrect link transport usage prediction. Highly similar plots are observed for all GNNs.}
    \label{results:q3:mae_vs_graph_size}
\end{figure}

\citet{Narayanan.2024} observed the existence of positive correlation between the graph size and the errors in the regression problem, meaning that the GNNs are not able to generalize well to larger networks. The same pattern continue to exist even in the fine grained classification problem, as depicted in Figure \ref{results:q3:mae_vs_graph_size}. On a different note, the improvements observed by increasing the training data is consistent with different GNN models. This implies that more training data is required for better predictions, as is the usual case for the deeper models. 

\begin{table}[!htb]
\small
\centering
\setstretch{0.7}
\captionsetup{justification=centering}
\caption{Evaluation metrics corresponding to the test dataset for fine grained classification with additional synthetic training data. GATv3 is in parentheses as it experienced overfitting.}
\label{results:q3:extra_data_put:results}
\vspace{0.1cm}
\begin{tabular}{@{}lcccc@{}}
\toprule
Model  & \multicolumn{2}{c}{Test Set}  \\[0.05cm]
 \hspace{2cm}  & $ R^2_{\, \ge10}$ & $MAE_{\, \ge10}$ \\ \midrule
Mean Regressor              & 0.00 & 402.6      \\
Random Forest               & 0.00  & 383.9    \\
MLP                         & 0.00  & 382.8  \\
GCNII - Regression Best  &  0.84  & 135.4 \\
GCNII - Buckets-nl54             &  0.79   &  136.4 \\ 
& & & & \\
\textbf{GCNII - Buckets-nl54  - extra data }&  \textbf{0.85}  & \textbf{118.5} \\ 
GATv3 - Buckets-nl54  - extra data &  (0.78)  & (144.9)  \\
\bottomrule
\end{tabular}
\end{table}

\section{Conclusions}
\label{sec:conclusions}
In this study, we developed advanced Graph Neural Network (GNN) architectures as surrogate models for strategic, long-term transport planning. Our work involved a comparative analysis between Graph Convolutional Networks (GCNs) and the more expressive Graph Attention Networks (GATs). To address the over-smoothing issue that limits the depth of GAT models, we introduced GATv3, a novel variant incorporating residual connections. Additionally, we proposed a fine-grained classification approach (along with a method to convert its output into precise values) that balances predictive stability with numerical precision, mitigating the challenges posed by complex regression tasks. To further improve model performance, we developed a synthetic data generator to expand training datasets without overfitting.

Our findings provide important insights into the performance of different model architectures, offering valuable guidance for future research and practical applications in the field. Notably, GATv3 demonstrated strong classification capabilities. The GATv3 formulation enables the development of deeper GAT models, making it particularly well-suited for classification-based transportation applications, where feature sharing across distant nodes is crucial. The results from the study also confirm that fine-grained classification is more stable with respect to model hyperparameters compared to regression tasks. Interestingly, in this task, GCNII emerged as the top performer when supplemented with additional synthetic training data, surpassing GATv3. All these suggest that problem type, task complexity, and dataset expansion play crucial roles in model performance. Despite these advancements, a detailed analysis reveals a linear relationship between network size and residual error, likely due to error propagation in message passing. As networks grow, errors accumulate and spread across nodes, necessitating further research into mitigating this effect in larger GNNs.

Overall, this study marks a significant step forward in refining GNN-based surrogate models for transport planning. The proposed GATv3 architecture and fine-grained classification framework offer promising new directions for broader transportation applications. On the one hand, GATv3 appears well-suited for classification-based transportation applications, such as section control and congestion warning systems, which require a higher degree of differentiation among neighboring links and feature sharing across distant nodes. On the other hand, GCNII is suited for the development of surrogates for travel demand modelling. Moving forward, future research should focus on reducing error propagation in deep GNNs, optimizing data augmentation strategies, and further refining GNN architectures for complex urban mobility scenarios.\\

\noindent\textbf{Declaration of interest:} None

\section*{Acknowledgements}
This research has been supported by European Union's Horizon Europe programme under grant agreement No. 101076963 [project PHOEBE (Predictive Approaches for Safer Urban Environment)].

\bibliography{references}
\end{document}